\documentclass[11pt]{article} 
\usepackage{times} %
\usepackage{acl2012}
\usepackage[dvipsnames]{color}
\usepackage[round]{natbib}
\usepackage{url}
\usepackage{xspace}
\usepackage{pdfsync}
\usepackage{fmtcount}
\newcommand{\ex}[1]{{`#1'}}
\newcommand{\hedge}[1]{{\em #1}}
\newcommand{\cut}[1]{} 
\newcommand{\trunc}[2]{#2} 

\title{Hedge detection as a lens on framing in the GMO debates: 
A  position paper\vspace*{-.03in}}

 \author{
\normalsize
Eunsol Choi$^*$, Chenhao Tan$^*$, Lillian Lee$^*$,  Cristian
   Danescu-Niculescu-Mizil$^*$
\and 
 Jennifer Spindel$^\dagger$\\ 
$^*$Department of Computer Science, $^\dagger$Department of  Plant Breeding and Genetics
\\
 Cornell University 
\\
 ec472@cornell.edu, chenhao$\vert$llee$\vert$cristian@cs.cornell.edu, jes462@cornell.edu}

\newcommand{\profsci}{prof-science\xspace}
\newcommand{\Profsci}{Prof-science\xspace}
\newcommand{\popsci}{pop-science\xspace}

\newcommand{\conlldata}{the CoNLL dataset\xspace}

\begin{document}
\maketitle

\begin{abstract}
 Understanding the ways in which participants in public discussions
 frame their arguments is important in understanding how public
 opinion is formed.  In this paper, we adopt the position that it is
 time for more computationally-oriented research on problems
 involving framing.  In the interests of furthering that goal, we
 propose the following specific, interesting and, we believe,
 relatively accessible question: In the controversy regarding the use
 of genetically-modified organisms (GMOs) in agriculture, do pro- and
 anti-GMO articles differ in whether they choose to adopt a more
 ``scientific'' tone?

Prior work on the rhetoric and sociology of science suggests that {\em
  hedging} may distinguish popular-science text from text written by
professional scientists for their colleagues.  We propose a detailed
approach to studying whether hedge detection can be used to
understanding scientific framing in the GMO debates, and provide
corpora to facilitate this study.  Some of our preliminary analyses
suggest that hedges occur less frequently in scientific discourse than
in popular text, a finding that contradicts prior assertions in the
literature.  We hope that our initial work and data will encourage
others to pursue this promising line of inquiry.

{\em Publication venue: ACL Workshop on Extra-Propositional Aspects of Meaning in Computational Linguistics, 2012}
\end{abstract}


\section{Introduction}
\label{sec:intro}

\subsection{Framing, ``scientific discourse'', and GMOs in the media}
The issue of {\em framing} \citep{Goffman:74a,Scheufele:99a,Benford+Snow:2000a} is of great importance in understanding how public opinion is formed.  In their {\em Annual Review
of Political Science} survey, \citet{Chong+Druckman:2007a} describe framing effects as occurring ``when (often small) changes in the presentation of
an issue or an event produce (sometimes large) changes of opinion'' (p. 104); as an example, they cite a study wherein respondents answered differently,
 when asked whether a hate group should be allowed to hold a rally, depending on whether the question was phrased as one of ``free speech'' or one of  ``risk of violence''.  

The genesis of our work is in a framing question
motivated by a relatively current political issue.
In media coverage of transgenic crops and 
the use of genetically modified organisms (GMOs) in food,
do pro-GMO vs. anti-GMO articles differ not just with respect to word choice,  but in adopting a 
more ``scientific'' discourse,
meaning 
the inclusion of 
more uncertainty
and
 fewer emotionally-laden words?
We view this as an interesting question from a text analysis perspective 
(with potential applications and implications that lie outside the scope of this article).

\subsection{Hedging as a sign of scientific discourse}
To obtain a computationally manageable characterization of ``scientific discourse'', we turned to studies of the culture and language of science, a body of work spanning fields ranging from sociology to applied
 linguistics
 to rhetoric and  communication
\citep{Gilbert:84a,Latour:87a,Latour+Woolgar:79a,Halliday+Martin:93a,Bazerman:1988,Fahnestock:2004a,Gross:90a}.

One characteristic that has drawn quite a bit of attention in such studies is
{\em hedging} \citep{Myers:89a,Hyland:98a,Lewin:98a,Salager-Meyer:2011a}.\footnote{In linguistics, hedging has been studied since the 1970s \citep{Lakoff:73a}.}
  \citet[pg. 1]{Hyland:98a} defines hedging as  the expression of ``tentativeness and possibility'' in communication, or, to put it another way, language corresponding to ``the writer withholding full commitment to statements'' (pg. 3).  He supplies many real-life examples from scientific research articles, including the following:
\begin{enumerate}
\item \ex{\hedge{It seems that} this group plays a critical role in orienting the carboxyl function} (emphasis Hyland's)
\item \ex{...\hedge{implies that} phytochrome A is also not necessary for normal photomorphogenesis, \hedge{at least under these irradiation conditions}}  (emphasis Hyland's)
\item  \ex{\hedge{We wish to suggest} a structure for the salt of deoxyribose nucleic acid (D.N.A.)} (emphasis added)\footnote{This example originates from Watson and Crick's landmark 1953 paper. Although the sentence is overtly tentative, did Watson and Crick truly intend to be polite and modest in their claims?  See \citet{Varttala:2001a} for a review of arguments regarding this question.}
\end{enumerate}
Several scholars have asserted the centrality of hedging in scientific and academic discourse, which corresponds nicely
to the 
notion
 of ``more uncertainty'' mentioned above.  \citet[p. 6]{Hyland:98a} writes, ``Despite a widely held belief that professional scientific writing is a series of impersonal statements of fact which add up to the truth, hedges are abundant in science and play a critical role in academic writing''.  Indeed, \citet[p. 13]{Myers:89a} claims that in scientific research articles, ``The hedging of claims is so common that a sentence that looks like a claim but has no hedging is probably not a statement of new knowledge''.\footnote{Note the inclusion of  the hedge ``probably''.}

Not only is understanding hedges important to understanding the rhetoric   and sociology of science, but hedge detection and analysis --- in the sense of identifying uncertain or uncertainly-sourced information \citep{Farkas+al:2010a} --- 
 has important applications to 
information extraction, broadly construed, 
and has thus become an active sub-area of natural-language processing.  For example,  the CoNLL 2010 Shared Task was devoted to this problem \citep{Farkas+al:2010a}.

Putting these two lines of research together, we 
see before us what 
appears to be
an interesting interdisciplinary and, at least in principle, straightforward research 
program:
relying on the aforementioned rhetoric analyses to presume that
hedging is a key characteristic of scientific discourse,
build a
hedge-detection system to computationally ascertain which proponents
in the GMO debate tend to use more hedges and thus, by 
presumption,
tend to adopt a more ``scientific''  frame.\footnote{However, this presumption that more
hedges characterize a more scientific discourse has been contested.
See section \ref{sec:hedging-distinguish} for discussion and section
\ref{sec:uncertain} for our empirical investigation.
}

\subsection{Contributions}
Our overarching goal in this paper is to convince
more researchers in NLP and computational linguistics to work on problems involving framing.
We try to do so by proposing a specific problem that may be relatively accessible.
 Despite the apparent difficulty in addressing such questions, we believe that progress can be made by drawing on observations drawn from previous literature across many fields, and integrating such work with movements in the computational community toward consideration of extra-propositional and pragmatic concerns.  We have thus intentionally tried to ``cover a lot of ground'', as one referee put it, in the introductory material just discussed.

Since framing problems are indeed difficult, we elected to narrow our scope in the hope of making some partial progress.
Our technical  goal here, at this workshop, where hedge detection is one of the most relevant topics to the broad questions we have raised,
is {\em not} to learn to classify texts as being pro- vs. anti-GMO, or as being scientific or not, per se.\footnote{
Several other groups have addressed the problem of trying to identify different sides or perspectives \citep{Lin+al:06a,Hardist+Boyd-Graber+Resnik:2010a,Klebanov:2010:VCI:1858842.1858889,Ahmed:2010:SIS:1870658.1870769}.
}
 Our focus is on whether hedging specifically, considered as a single feature, is correlated with these different document classes, because of the previous research attention that has been devoted to hedging in particular
and because of hedging being one of the topics of this workshop.
The point of this paper is thus not to compare the efficacy of hedging features with other types,
such as 
bag-of-words features. 
Of course, to do so is an important and interesting direction for future work.

In the end, we were not able to achieve satisfactory results even with
respect to our narrowed goal.  However, we believe that other
researchers may be able to follow the plan of attack we outline below,
and perhaps use the data we are releasing, in order to  achieve our
goal.  We would welcome hearing the results of 
other 
people's
efforts.

\section{How should we
test
whether hedging 
distinguishes
scientific text?}
\label{sec:hedging-distinguish}
One very important point that we have not yet addressed is:
While the literature agrees on the importance of hedging in scientific text, 
 the {\em relative
  degree} of hedging in scientific vs. non-scientific text is a matter
of 
debate.

On the one side, we have assertions like those of
\citet{Fahnestock:86a}, who shows in a clever, albeit small-scale,
study involving parallel texts that when scientific observations pass into popular accounts,
changes include ``removing hedges 
 ... thus
conferring greater certainty on the reported facts'' (pg. 275).  Similarly, \citet{Juanillo:2001a} refers to a
shift from a forensic style to a ``celebratory'' style when scientific research becomes publicized, and credits
\citet{Brown:98a} with noting that ``celebratory scientific discourses
tend to pay less attention to caveats, contradictory evidence, and
qualifications that are highlighted in forensic or empiricist
discourses. By downplaying scientific uncertainty, it [sic] alludes to greater certainty of scientific results for public consumption'' \citep[p. 42]{Juanillo:2001a}.  

However, others have contested claims that the popularization process involves simplification, distortion, hype, and dumbing down, as \citet{Myers:2003a} colorfully puts it; he provides a critique of the relevant literature.   \citet{Varttala:99a} ran a corpus analysis in which hedging was found not just in professional medical articles, but was also ``typical of popular scientific articles dealing with similar topics'' (p. 195).   Moreover, significant variation in use of hedging has been found across disciplines and authors' native language; see \citet{Salager-Meyer:2011a} or \citet{Varttala:2001a} for a review.

To the best of our knowledge, there 
have been no large-scale empirical studies
validating the hypothesis
that hedges appear more 
or less 
frequently
 in scientific discourse.

\paragraph{Proposed procedure}
Given the above, our {\bf first step} must be to determine whether hedges
are more or less prominent in 
``professional scientific'' (henceforth
``{\em \profsci'}') vs. ``public science'' (henceforth ``{\em \popsci}'')
discussions of GMOs.
Of course, 
for a large-scale study,
finding hedges requires developing and training an effective hedge
detection algorithm.

 If the first step shows that hedges can indeed be
used to
effectively 
distinguish 
\profsci vs. \popsci
discourse on GMOs,
then the {\bf second
step} is to examine whether the use of hedging in pro-GMO articles
follows our inferred ``scientific'' occurrence patterns 
to a greater extent
than the 
hedging in anti-GMO articles.

However, 
as our hedge classifier trained on
\conlldata
did not perform reliably on the different domain
of \profsci vs. \popsci discussions of GMOs,
we focus the main content of this paper on the first step. We describe data collection for the second step
in the appendix.


\section{Data}

\begin{table*}[htb!]
{\small
\begin{center}
\addtocounter{footnote}{1}
\begin{tabular}{|l|c|r|r|r|r|}
\hline
Dataset & Doc type & \# docs & \# sentences& Avg sentence length &
Flesch reading ease\\
\hline \hline
\multicolumn{6}{|c|}{\Profsci/\popsci corpus} \\ \hline
WOS & abstracts & 648 & 5596 & 22.35 & 23.39\\ 
LEXIS & (short) articles & 928 & 36795 & 24.92
& 45.78\\
\hline
\hline
\multicolumn{6}{|c|}{Hedge-detection corpora} \\ \hline
Bio (train) & abstracts, articles & 1273, 9 
& 14541 (18\% uncertain) & 29.97 & 20.77 \\
Bio (eval) & articles & 15 & 5003 (16\% uncertain)& 31.30 & 30.49\\
Wiki (train) & paragraphs &  2186 & 11111 (22\% uncertain) & 23.07 & 35.23\\
Wiki  (eval) & paragraphs &  2346 & 9634 (23\% uncertain) & 20.82 & 31.71\\ \hline
\end{tabular}
\end{center}
}
\caption{\label{tab:corpora} 
Basic descriptive statistics for the main corpora we worked with.  We
created the first two.   Higher Flesch scores indicate text that is
easier to read.
} 
\end{table*}

To accomplish the first step of our proposed procedure outlined above, we
first
constructed
 a \profsci/\popsci corpus by pulling text from Web of
Science for \profsci examples and from LexisNexis for \popsci
examples, as described in Section \ref{sec:getsci}.   Our corpus will
be posted online at \url{https://confluence.cornell.edu/display/llresearch/HedgingFramingGMOs}.

As noted above, computing the degree of hedging in the aforementioned corpus requires
access to a hedge-detection algorithm.  We took a supervised approach,
taking advantage of  the availability of the CoNLL 2010
hedge-detection training and 
evaluation
corpora, described in Section \ref{sec:conlldata}

\subsection{\Profsci/\popsci data: LEXIS and WOS}
\label{sec:getsci}

As mentioned 
previously, 
a corpus of \profsci and \popsci
articles 
is required
 to ascertain whether hedges are more prevalent in
one 
or the other of these 
two
writing styles.  Since our ultimate
goal is to look at discourse related to 
GMOs,
we
restrict our attention to documents on 
this
topic.

Thomson Reuter's Web of Science (WOS), a database of scientific
journal and conference
articles,
 was used as a source of \profsci samples. We chose to collect
abstracts, rather than full scientific articles, because intuition
suggests that the language in abstracts is more high-level than
that in the bodies of papers, and
thus more similar to the language
one would see in a public debate on GMOs. To 
select for on-topic
abstracts, we used the phrase ``transgenic foods" as
a
 search keyword and
discarded 
results containing any of a hand-selected list of
off-topic filtering terms (e.g., ``mice'' or ``rats'').  
We then made use of domain expertise
to manually
remove off-topic texts. 
The process
yielded 648 documents 
for a total of 
5596 sentences.

Our source of \popsci articles was LexisNexis (LEXIS).
On-topic
documents were collected from US newspapers using the search keywords
``genetically modified foods" or ``transgenic crops" and then
imposing the additional requirement that at least two terms on a
hand-selected list\footnote{
The term list:
GMO, GM, GE, genetically modified, genetic
  modification, modified, modification, genetic engineering,
  engineered, bioengineered, franken, transgenic, spliced, G.M.O.,
  tweaked, manipulated, engineering, pharming, aquaculture.}
be
present in each document.  After the removal of duplicates and
texts containing more than 2000 words to
delete excessively long articles,
our final \popsci subcorpus was composed of 928
documents.

\subsection{CoNLL hedge-detection training data \footnote{\url{http://www.inf.u-szeged.hu/rgai/conll2010st/}}}
\label{sec:conlldata}
As described in \citet{Farkas+al:2010a}, the motivation behind the CoNLL 2010 shared task is that
``distinguishing factual and uncertain information in texts is of
essential importance in information extraction''.
As ``uncertainty detection is extremely important for biomedical
information extraction'', one component of the dataset is biological
abstracts and full articles from the
 BioScope
corpus
 (Bio). 
Meanwhile, the chief editors of Wikipedia have drawn the attention of
the public to 
specific markers of uncertainty
known as
weasel
words\footnote{\url{http://en.wikipedia.org/wiki/Weasel_word}}:
they are words or phrases
``aimed at creating an impression that
something specific and meaningful has been said'', when,
 in fact,
``only a
vague or ambiguous claim, or even a refutation,
 has been communicated''.
An example is ``It has been claimed that ...'':
the claimant has not been identified, so the source of the claim
cannot be verified.
Thus,
 another part of the dataset is 
a set of
Wikipedia articles 
(Wiki)
annotated with weasel-word information.
We view
the combined Bio+Wiki corpus
(henceforth \conlldata) 
as 
valuable 
for developing 
hedge
detectors,
and we
attempt
to study whether classifiers trained on this data
can be generalized to other datasets.

\subsection{Comparison}
Table \ref{tab:corpora} gives the basic statistics 
on 
the main datasets we worked with.
Though WOS and LEXIS differ in the total number of
sentences, the average sentence length is 
similar.
The average sentence length in Bio is 
longer than that in Wiki.
The 
articles in WOS 
are 
markedly
more difficult to read than 
the articles in 
LEXIS according to
Flesch reading ease \citep{Kincaid:1975}.


\section{Hedging to distinguish scientific text: 
Initial annotation}
\label{sec:hedge}

As noted in Section \ref{sec:intro}, it is not
a priori 
 clear 
whether
hedging 
distinguishes 
scientific text 
or that
more hedges correspond to 
a more ``scientific'' discourse. 
To get an initial feeling for how 
frequently hedges occur in WOS and LEXIS, we
hand-annotated a sample of sentences from each.
In Section \ref{sec:annotate}, 
we explain 
the 
annotation policy of
the
 CoNLL 2010 Shared Task and 
our own annotation 
method for 
WOS and 
LEXIS.
After that, we move forward 
in Section \ref{sec:uncertain} to compare the
percentage of uncertain sentences in 
\profsci vs. \popsci
text
on this small hand-labeled sample, and gain some evidence that there is indeed a difference in hedge occurrence rates, although, perhaps surprisingly,  there seem to be more hedges in the {\em \popsci} texts.

As a side benefit, we subsequently use the hand-labeled sample we produce 
to investigate the accuracy of an automatic hedge detector in the WOS+LEXIS
domain; more on this in Section \ref{sec:exp}.

\subsection{Uncertainty annotation}
\label{sec:annotate}

\paragraph{CoNLL 2010 Shared Task annotation policy} As described in
\citet[pg. 4]{Farkas+al:2010a}, the data annotation polices for the
CoNLL 2010 Shared Task were
 that ``sentences containing at least one
cue were considered as uncertain, while sentences with no cues were
considered as factual'', where a cue is a linguistic marker that {\em
  in context} 
indicates uncertainty.  A straightforward example of a
sentence marked ``uncertain'' in the Shared Task is \ex{Mild bladder
  wall thickening \hedge{raises the question of} cystitis.} 
 The annotated cues are not necessarily general,
particularly in Wiki, where some of the marked
cues are as specific as  \ex{\hedge{some of schumann's
   best choral writing}},  \ex{\hedge{people of the jewish tradition}},
or
\ex{\hedge{certain leisure or cultural activities}}. 

Note that ``uncertainty'' in the Shared Task definition also encompassed phrasing that ``creates an impression that something important has been said, but what is really communicated is vague, misleading, evasive or ambiguous ... [offering] an opinion without any backup or source''.  An example of such a sentence, drawn from Wikipedia and marked ``uncertain'' in the Shared Task, is \ex{Some people claim that this results in a better taste than that of other diet colas (most of which are sweetened with aspartame
alone).}; \citet{Farkas+al:2010a} write, ``The ... sentence does not specify the source of the information, it is just the vague term \ex{some people} that refers to the holder of this opinion''.

\paragraph{Our annotation policy} 
We hand-annotated 200 
randomly-sampled
 sentences,
half from WOS and half from LEXIS\footnote{ 
We took steps to 
attempt to hide from the annotators any explicit clues 
as to the source of individual sentences: the subset of authors who did the annotation were not those that collected the data, and the annotators were presented the sentences in random order.},
to gauge the frequency
with which
hedges occur in each corpus.
Two annotators
each 
 followed the rules 
of the 
CoNLL 2010 Shared Task to label sentences as certain, 
uncertain, or 
not a proper sentence.\footnote{
The last label was added because of 
a few errors in scraping the data.
}
The annotators agreed on 153 proper sentences of the 200
sentences (75 from WOS and 78 from
LEXIS).
Cohen's Kappa \citep{fleiss:81} was \trunc{0.668}{0.67} on the annotation, 
which 
means that the consistency between
the
 two annotators 
was fair or good. 
However, there 
were some interesting cases 
where the two annotators could not agree.
For example, in the 
sentence
\ex{Cassava is the staple food of tropical Africa and its production,
 averaged over 24 countries, has increased more than threefold from 1980
 to 2005 
\mbox{...
}},
 one of the 
annotators
 believed that ``more than'' 
made the sentence uncertain. These
 borderline cases 
indicate
that 
the definition of
hedging should be carefully 
delineated
in
future
 studies.

\subsection{Percentages of uncertain sentences}
\label{sec:uncertain}

\begin{table}
\begin{center}
\begin{tabular}{|l|r|}
\hline
Dataset & \% of uncertain sentences  \\
\hline
WOS &  (estimated from 75-sentence sample) \trunc{20.00}{20} \\
LEXIS &  (estimated  from 78-sentence sample) \trunc{28.21}{28}  \\
Bio & 17
 \\
Wiki & 23
 \\
\hline
\end{tabular}
\caption{\label{tab:uncertainty}  
Percentages of uncertain sentences.
  }
\end{center}
\end{table}

To validate the hypothesis that 
\profsci articles
contain
more hedges, we 
computed
the percentage of uncertain sentences in our
labeled data. 
As shown in Table \ref{tab:uncertainty}, 
we
observed a trend 
contradicting
 earlier studies.
Uncertain sentences 
were 
more frequent in LEXIS than in WOS, though
the 
difference
was not
statistically significant\footnote{
Throughout, ``statistical significance'' refers to the student t-test
with $p < .05$.}
(perhaps not surprising given the small sample size).
The same trend was seen
in \conlldata:
there, too,
the
percentage of uncertain sentences 
was
significantly
smaller in 
Bio (\profsci articles)
than in Wiki.
In order to make
a stronger argument about \profsci vs \popsci, 
however,
more annotation on the WOS and LEXIS datasets is needed.


\newcommand{\best}[1]{{\bf #1}}\newcommand{\secondbest}[1]{{\em #1}}

\section{Experiments}
\label{sec:exp}

As 
stated
in Section \ref{sec:intro}, 
our proposal requires developing an 
effective hedge detection algorithm. 
Our approach for the preliminary work described in this paper is to re-implement Georgescul's 
\citeyearpar{Georgescul:2010a}
algorithm; the experimental results on the Bio+Wiki domain, given in  Section \ref{sec:method}, are encouraging.
Then we
use this method to 
attempt to
validate
(at a larger scale than in our manual pilot annotation)
whether hedges can 
be used to distinguish 
between 
\profsci 
and \popsci
discourse on GMOs.
Unfortunately, our results, given in Section \ref{sec:wosresults}, are inconclusive, since our trained model could not achieve satisfactory automatic hedge-detection accuracy on the WOS+LEXIS domain.

\subsection{Method}
\label{sec:method}

We adopted the method of 
\citet{Georgescul:2010a}:
Support Vector Machine
classification based on 
a
Gaussian Radial Basis kernel function
\citep{Vapnik:98,Fan:2005},
employing n-grams from annotated cue phrases as features, as described in more detail below.
This method
achieved the
top performance in the CoNLL 2010 Wikipedia hedge-detection task
\citep{Farkas+al:2010a},
and SVMs 
have
been 
proven
 effective 
for many
 different applications.
We used
the
 LIBSVM toolkit in our 
experiments\footnote{\url{http://www.csie.ntu.edu.tw/~cjlin/libsvm/}}.

As described in Section \ref{sec:conlldata}, there are two separate
datasets 
in \conlldata .
We experimented on them separately (Bio, Wiki).
Also,
to make our classifier
more generalizable 
to
different datasets, we also trained models based on the two datasets
combined (Bio+Wiki).
As for features, we took advantage of the observation in
\citet{Georgescul:2010a} that 
the
bag-of-words model 
does not work well for
this task. 
We used different 
sets
of features based on hedge 
{\em cue words}
that have been annotated
as part of the CoNLL dataset distribution\footnote{
For the Bio model, we
used cues extracted from
 Bio. Likewise, the Wiki model used cues from Wiki, and the Bio+Wiki model
 used cues from Bio+Wiki.}.
The basic feature set was the frequency of each hedge cue 
word
 from the
training corpus after removing stop words
and 
punctuation
 and 
transforming words to
lowercase.
Then, we extracted unigrams, bigrams and trigrams from 
each
hedge cue
phrase.
Table \ref{tab:cues} shows the number of features in different
settings.
Notice that there are many more features in Wiki.
As mentioned above, 
in Wiki, 
some
cues are as specific as
  \ex{some of schumann's
   best choral writing}, 
 \ex{people of the jewish tradition}, 
or
\ex{ certain leisure or cultural activities}. Taking 
n-grams
from 
such
specific cues can 
cause some sentences to be classified incorrectly.

\begin{table}[th]
\begin{center}
\begin{tabular}{|l|r|}
\hline
Feature source & \#features\\
\hline
Bio & 220\\
Bio (cues + bigram + trigram) & 340 \\
Wiki &3740 \\
Wiki (cues + bigram + trigram) & 10603 \\
 \hline
\end{tabular}
\end{center}
\caption{\label{tab:cues} Number of features. 
}
\end{table}

\begin{table}[th]
\begin{center}
\begin{tabular}{| c | c || c | c | c |}
\hline
\multicolumn{5}{|c|}{Best 
cross-validation performance}\\
\hline
Dataset & $(C, \gamma)$  & P & R & F \\
\hline
Bio  & $(40, 2^{-3})$ & \best{\trunc{0.840}{84.0}} & \best{\trunc{0.920}{92.0}} & \best{\trunc{0.878}{87.8}}  \\
Wiki & $(30, 2^{-6})$ & \trunc{0.640}{64.0} &  \trunc{0.763}{76.3} & \trunc{0.696}{69.6} \\
Bio+Wiki & $(10, 2^{-4})$ &  \secondbest{\trunc{0.667}{66.7}} & \secondbest{\trunc{0.783}{78.3}} & \secondbest{\trunc{0.720}{72.0}}  \\
\hline
\end{tabular}
\normalsize
\end{center}
\caption{\label{tab:perfCV} 
Best 5-fold cross-validation performance for Bio and/or Wiki after parameter tuning.  
As a reminder, we repeat that our intended final test set is the WOS+LEXIS corpus, which is disjoint from Bio+Wiki.
}
\end{table}

We adopted several 
techniques
from \citet{Georgescul:2010a} to
optimize performance through 
cross validation.
Specifically, we tried different 
combinations
 of feature sets
(the cue phrases themselves,
cues + unigram, cues + bigram, cues + trigram, cues + unigram +
bigram + trigram, cues + bigram + trigram).
We tuned the width of 
the
RBF kernel ($\gamma$) and the regularization
parameter ($C$) via grid search over the following range of values: $\{2^{-9}, 2^{-8}, 2^{-7}, \ldots, 2^{4}\}$ 
for $\gamma$, $\{1,10,20,30,\ldots,150\}$ for $C$. 
We also tried different weighting strategies for negative and
positive classes 
(i.e., either
proportional to the number of 
positive
instances, or uniform).
We performed 5-fold 
cross validation 
for each possible combination of
experimental settings on 
the
three datasets (Bio, Wiki, Bio+Wiki).

Table \ref{tab:perfCV} shows the best performance on 
all 
three datasets
and the corresponding parameters. In 
the three datasets,
cue+bigram+trigram 
provided the best performance, 
and the weighted model consistently produced superior results to 
the 
uniform
model. 
The F1 measure for
Bio was \trunc{0.878}{87.8}, which
was satisfactory, while
the F1 results for
Wiki 
were \trunc{0.696}{69.6}, 
which 
were the worst of
all
the datasets. This resonates with our observation that the task on Wikipedia is
more subtly defined 
and thus requires 
a more sophisticated approach than counting the
occurrences of bigrams and trigrams.

\subsection{Results on WOS+LEXIS}
\label{sec:wosresults}

\begin{table}[t]
\begin{center}
{\small
\begin{tabular}{| c | c || c | c | c |}
\hline
Evaluation set& Model & P & R & F \\
\hline
WOS+LEXIS & Bio & \best{\trunc{0.543}{54}}  & \trunc{0.676}{68} &  \best{\trunc{0.602}{60}} \\
WOS+LEXIS & Wiki & \trunc{0.385}{38} & \trunc{0.541}{54} & \trunc{0.449}{45} \\
WOS+LEXIS & Bio+Wiki & \trunc{0.212}{21}  & \best{\trunc{0.933}{93}} &  \trunc{0.345}{34} \\
\hline
\hline
\multicolumn{5}{|c|}{Sub-corpus performance of the model based on Bio}\\
\hline
WOS  & Bio &{\trunc{0.579}{58}} & {\trunc{0.733}{73}} &  {\trunc{0.647}{65}} \\
LEXIS & Bio & \trunc{0.519}{52}  & \trunc{0.636}{64} &  \trunc{0.571}{57} \\
\hline
\end{tabular}
}
\caption{\label{tab:perfWL} The upper part shows the performance on
  WOS and LEXIS based on models trained 
on 
\conlldata. The lower part
gives the sub-corpus results
for Bio,
  which 
provided
the best performance on 
the full
WOS+LEXIS
corpus.}
\end{center}
\end{table}

\begin{table}[t]
\begin{center}
\begin{tabular}{|c|c||c|c|c|}
\hline
Evaluation set & $(C,\gamma)$ & P & R & F \\
\hline
WOS + LEXIS & $(50, 2^{-9})$ & \trunc{0.676}{68} & \trunc{0.621}{62} & \trunc{0.648}{65}\\
WOS & $(50, 2^{-9})$ &\trunc{0.846}{85} & \trunc{0.733}{73} & \trunc{0.786}{79}\\
LEXIS & $(50, 2^{-9})$ & \trunc{0.571}{57} & \trunc{0.545}{54} & \trunc{0.558}{56}\\
\hline
\end{tabular}
\end{center}
\caption{\label{tab:tunedWL} Best performance after parameter tuning
  based on the
 153 labeled WOS+LEXIS sentences;
this gives some idea of the upper-bound potential of our Georgescul-based method.
 The training set is 
Bio,
which gave the best performance in Table \ref{tab:perfWL}.}
\end{table}

Next, we evaluated
whether 
our
best
 classifier trained on 
\conlldata can be generalized 
to
other
datasets, in particular, 
the
WOS and LEXIS 
corpus.
Performance was measured on the 153 sentences on which our annotators agreed, a dataset that was
 introduced in
Section \ref{sec:annotate}. 
Table \ref{tab:perfWL} shows how the best models trained 
on Bio,
Wiki, and Bio+Wiki,
respectively,
performed 
on the
 153 labeled sentences.
First, we can see
that the performance 
degraded
significantly compared to the performance
for in-domain 
cross validation.
Second,
of the three different models, Bio
showed
the best performance. Bio+Wiki 
gave
the worst performance,
which hints 
that combining two datasets 
 and cue words
may not be
a
 promising
strategy:
although Bio+Wiki shows 
very good recall,
this
can be attributed to
its
larger feature set, 
which contains
 all available cues
and perhaps as a result has a very high false-positive rate. 
We further
investigated and compared performance on LEXIS and WOS 
for
the best model
 (Bio). 
Not surprisingly, 
our classifier works better in WOS than in LEXIS.

It is clear that there 
exist
domain
 differences between 
\conlldata
and WOS+LEXIS. 
To better understand the poor
cross-domain
 performance of the classifier, we
tuned
another model based on the performance on the 153 labeled
sentences using Bio as training data. 
As we can see in Table \ref{tab:tunedWL}, the performance
on WOS 
improved significantly, while the performance on LEXIS 
decreased.
This is probably caused by the fact that WOS is a
collection of scientific paper abstracts, which is more similar to
the training corpus than LEXIS, which is a collection
 of news media articles\footnote{The Wiki model performed better on
   LEXIS than on WOS. Though the performance 
was not good, this result further reinforces the possibility of a
domain-dependence problem.
 }. 
Also, LEXIS articles are hard to classify even with 
the
tuned model,
 which
challenges the effectiveness of 
a cue-words frequency approach 
beyond professional scientific texts.
Indeed, 
the
 simplicity of our 
reimplementation of Georgescul's
algorithm 
seems to 
cause
longer sentences 
to be classified as uncertain,
because 
cue 
phrases (or n-grams extracted from cue phrases) are 
more
likely to appear in 
lengthier sentences. 
Analysis of
the best performing model 
shows
that the 
false-positive sentences are significantly 
longer than the 
false-negative
ones.\footnote{Average length of true positive sentences : 28.6, false
positive sentences 35.09, false negative sentences: 22.0.}

\begin{table}[htb!]
\begin{center}
\begin{tabular}{| c | c | c |}
\hline
Dataset & Model & \% classified uncertain  \\
\hline
WOS & Bio & \trunc{16.33}{16} \\
LEXIS & Bio &  \trunc{18.67}{19} \\
WOS & Tuned &\trunc{15.40}{15}  \\
LEXIS & Tuned &   \trunc{14.24}{14} \\ 
\hline
\end{tabular}
\end{center}
\caption{\label{tab:uncertainWL} 
For completeness, we report here the
percentage of uncertain sentences in
  WOS and LEXIS 
according to our trained classifiers, although we regard these results as unreliable
since those classifiers have low accuracy. 
Bio refers to the best model trained on Bio only in
  Section \ref{sec:method}, while Tuned refers to the model in Table
  \ref{tab:tunedWL} that is
  tuned based on 
the
153 labeled sentences in WOS+LEXIS.}
\end{table}

While the cross-domain results were not reliable, we produced
preliminary results on whether there 
exist fewer hedges in scientific
text. We can see that the relative 
difference in certain/uncertain ratios predicted
by the two
different models (Bio, Tuned) are different
in Table \ref{tab:uncertainWL}. 
 In the tuned model, the difference between LEXIS and WOS in terms of
 the percentage of uncertain sentences was 
not statistically significant, 
while in
the
 Bio model, their difference was statistically significant.
Since the performance of our hedge classifier on 
the
153
  hand-annotated 
WOS+LEXIS
sentences
was not reliable,
though,
 we 
must abstain from 
making
  conclusive statements here.


\section{Conclusion and future work}

In this position paper, we
advocated that researchers apply hedge detection not only to the
classic motivation of
information-extraction problems, but also to questions of how public
opinion forms.  We
proposed a particular problem in how
participants in debates frame their arguments.
Specifically, 
we asked
whether pro-GMO and anti-GMO articles differ in adopting 
a more ``scientific'' discourse.
Inspired by earlier studies in social sciences relating
  hedging to 
texts aimed at professional scientists,
we 
proposed
addressing the question
with 
automatic hedge
detection as a first step. 
To develop
 a hedge classifier, we
took advantage of 
\conlldata
and a small
annotated WOS and LEXIS dataset. 
Our preliminary results show there
may
 exist a gap
which indicates
 that hedging may,
in fact,
 distinguish \profsci and \popsci documents. 
In fact,
this computational analysis suggests 
the possibility
that hedges occur less 
frequently
 in scientific prose, which 
contradicts 
several
prior assertions in
the literature. 

To confirm the argument 
that {\popsci} tends to use more hedging than
{\profsci}, we need a hedge classifier 
that
performs more
reliably in 
the
WOS and LEXIS dataset than ours
does. 
An interesting
research direction would be to develop
 transfer-learning techniques to
generalize 
hedge classifiers
for
 different datasets, or to develop
a
general hedge classifier relatively robust to domain
differences. 
In either case, more annotated data on WOS and LEXIS is needed for
better evaluation or training.

Another strategy would be to bypass the first step, in which we
determine whether hedges are more or less prominent in scientific
discourse, and proceed directly to labeling and hedge-detection in
pro-GMO and anti-GMO texts. However, this will not answer the question
of whether 
advocates in debates other than on GMO-related topics
employ a more scientific discourse.
Nonetheless, to aid those who wish to pursue this alternate strategy, 
we have collected two sets of opinionated articles on GMO (pro- and
anti-); see appendix for more details.

{


\paragraph{Acknowledgments} 
We thank
Daniel Hopkins and 
Bonnie Webber  
for reference suggestions, 
and the anonymous reviewers
for helpful and thoughtful comments.
This paper is based upon work supported in part by US NSF grants IIS-0910664 and  IIS-1016099, a US NSF graduate fellowship to JS, Google, and Yahoo!

}


\section{Appendix: pro- vs. anti-GMO dataset}
\label{sec:appendix}

Here, we describe the pro- vs. anti-GMO dataset we collected, in the
hopes that this dataset may prove helpful in future research regarding
the GMO debates, even though
we did not use the corpus in the project described in this paper.

The second step of our 
overall procedure outlined in the 
introduction
--- 
that step being
to examine whether the
use of hedging in pro-GMO articles corresponds with our inferred
``scientific'' occurrence patterns more than that in anti-GMO articles
---
requires a collection of 
opinionated articles on GMOs. 
Our first attempt to use news media
articles
(LEXIS) was unsatisfying, as we found many articles attempt to 
maintain
a
neutral position. This led us to collect documents from more strongly opinionated
organizational websites such as 
Greenpeace (anti-GMO), Non GMO
Project (anti-GMO), 
or
Why
 Biotechnology (pro-GMO). Articles were collected from 20
 pro-GMO and 20 anti-GMO organizational web sites.

After the initial collection of data,
 near-duplicates and irrelevant articles were filtered through clustering, keyword searches and
 distance between word vectors at 
the
document level. We have collected
 762 ``anti'' documents and 671 ``pro'' documents. 
We reduced this to a 404 ``pro'' and 404 ``con'' set as follows.
Each 
retained
``document''
consists of only the
first 200 words 
after 
excluding 
the
first 50 words
of documents containing over 280 words.
This was done
to 
avoid 
irrelevant 
sections
such as {\it 
Educators have permission to reprint articles for
classroom use; other users, please contact editor@actionbioscience.org
for reprint permission. See reprint policy.} 

The data will be posted online at \url{https://confluence.cornell.edu/display/llresearch/HedgingFramingGMOs}.


\begin{thebibliography}{32}
\providecommand{\natexlab}[1]{#1}
\providecommand{\url}[1]{\texttt{#1}}
\expandafter\ifx\csname urlstyle\endcsname\relax
  \providecommand{\doi}[1]{doi: #1}\else
  \providecommand{\doi}{doi: \begingroup \urlstyle{rm}\Url}\fi

\bibitem[Ahmed and Xing(2010)]{Ahmed:2010:SIS:1870658.1870769}
Amr Ahmed and Eric~P Xing.
\newblock {Staying informed: supervised and semi-supervised multi-view topical
  analysis of ideological perspective}.
\newblock In \emph{EMNLP}, pages 1140--1150, 2010.

\bibitem[Bazerman(1988)]{Bazerman:1988}
Charles Bazerman.
\newblock \emph{Shaping Written Knowledge: The Genre and Activity of the
  Experimental Article in Science}.
\newblock University of Wisconsin Press, Madison, Wis., 1988.

\bibitem[{Beigman Klebanov} et~al.(2010){Beigman Klebanov}, Beigman, and
  Diermeier]{Klebanov:2010:VCI:1858842.1858889}
Beata {Beigman Klebanov}, Eyal Beigman, and Daniel Diermeier.
\newblock Vocabulary choice as an indicator of perspective.
\newblock In \emph{ACL Short Papers}, pages 253--257, Stroudsburg, PA, USA,
  2010. Association for Computational Linguistics.

\bibitem[Benford and Snow(2000)]{Benford+Snow:2000a}
Robert~D. Benford and David~A. Snow.
\newblock {Framing processes and social movements: An overview and assessment}.
\newblock \emph{Annual Review of Sociology}, 26:\penalty0 611--639, 2000.

\bibitem[Brown(1998)]{Brown:98a}
Richard~Harvey Brown.
\newblock \emph{Toward a democratic science: Scientific narration and civic
  communication}.
\newblock Yale University Press, New Haven, 1998.

\bibitem[Chong and Druckman(2007)]{Chong+Druckman:2007a}
Dennis Chong and James~N. Druckman.
\newblock {Framing theory}.
\newblock \emph{Annual Review of Political Science}, 10:\penalty0 103--126,
  2007.

\bibitem[Fahnestock(1986)]{Fahnestock:86a}
Jeanne Fahnestock.
\newblock {Accommodating Science}.
\newblock \emph{Written Communication}, 3\penalty0 (3):\penalty0 275--296,
  1986.

\bibitem[Fahnestock(2004)]{Fahnestock:2004a}
Jeanne Fahnestock.
\newblock {Preserving the figure: Consistency in the presentation of scientific
  arguments}.
\newblock \emph{Written Communication}, 21\penalty0 (1):\penalty0 6--31, 2004.

\bibitem[Fan et~al.(2005)Fan, Chen, and Lin]{Fan:2005}
Rong-En Fan, Pai-Hsuen Chen, and Chih-Jen Lin.
\newblock Working set selection using second order information for training
  support vector machines.
\newblock \emph{JMLR}, 6:\penalty0 1889--1918, December 2005.
\newblock ISSN 1532-4435.

\bibitem[Farkas et~al.(2010)Farkas, Vincze, M\'ora, Csirik, and
  Szarvas]{Farkas+al:2010a}
Rich\'ard Farkas, Veronika Vincze, Gy\"orgy M\'ora, J\'anos Csirik, and
  Gy\"orgy Szarvas.
\newblock {The CoNLL-2010 shared task: Learning to detect hedges and their
  scope in natural language text}.
\newblock In \emph{CoNLL---Shared Task}, pages 1--12, 2010.

\bibitem[Fleiss(1981)]{fleiss:81}
Joseph~L. Fleiss.
\newblock \emph{{Statistical Methods for Rates and Proportions}}.
\newblock Wiley series in probability and mathematical statistics. John Wiley
  \& Sons, New York, second edition, 1981.

\bibitem[Georgescul(2010)]{Georgescul:2010a}
Maria Georgescul.
\newblock {A hedgehop over a max-margin framework using hedge cues}.
\newblock In \emph{CONLL---Shared-Task}, pages 26--31, 2010.

\bibitem[Gilbert and Mulkay(1984)]{Gilbert:84a}
G.~Nigel Gilbert and Michael~Joseph Mulkay.
\newblock \emph{Opening Pandora's box: A sociological analysis of scientists'
  discourse}.
\newblock CUP Archive, 1984.

\bibitem[Goffman(1974)]{Goffman:74a}
Erving Goffman.
\newblock \emph{Frame analysis: An essay on the organization of experience}.
\newblock Harvard University Press, 1974.

\bibitem[Gross(1990)]{Gross:90a}
Alan~G. Gross.
\newblock \emph{The rhetoric of science}.
\newblock Harvard University Press, Cambridge, Mass., 1990.

\bibitem[Halliday and Martin(1993)]{Halliday+Martin:93a}
Michael Alexander~Kirkwood Halliday and James~R. Martin.
\newblock \emph{Writing science: Literacy and discursive power}.
\newblock Psychology Press, London [u.a.], 1993.

\bibitem[Hardisty et~al.(2010)Hardisty, Boyd-Graber, and
  Resnik]{Hardist+Boyd-Graber+Resnik:2010a}
Eric~A Hardisty, Jordan Boyd-Graber, and Philip Resnik.
\newblock {Modeling perspective using adaptor grammars}.
\newblock In \emph{EMNLP}, pages 284--292, 2010.

\bibitem[Hyland(1998)]{Hyland:98a}
Ken Hyland.
\newblock \emph{Hedging in scientific research articles}.
\newblock John Benjamins Pub. Co., Amsterdam; Philadelphia, 1998.

\bibitem[{Juanillo, Jr.}(2001)]{Juanillo:2001a}
Napoleon~K. {Juanillo, Jr.}
\newblock {Frames for Public Discourse on Biotechnology}.
\newblock In \emph{Genetically Modified Food and the Consumer: Proceedings of
  the 13th meeting of the National Agricultural Biotechnology Council}, pages
  39--50, 2001.

\bibitem[Kincaid et~al.(1975)Kincaid, Fishburne, Rogers, and
  Chissom]{Kincaid:1975}
J.~Peter Kincaid, Robert~P. Fishburne, Richard~L. Rogers, and Brad~S. Chissom.
\newblock {Derivation of new readability formulas for navy enlisted personnel.}
\newblock Technical report, National Technical Information Service,
  Springfield, Virginia, February 1975.

\bibitem[Lakoff(1973)]{Lakoff:73a}
George Lakoff.
\newblock {Hedges: A study in meaning criteria and the logic of fuzzy
  concepts}.
\newblock \emph{Journal of Philosophical Logic}, 2\penalty0 (4):\penalty0
  458--508, 1973.

\bibitem[Latour(1987)]{Latour:87a}
Bruno Latour.
\newblock \emph{Science in action: How to follow scientists and engineers
  through society}.
\newblock Harvard University Press, Cambridge, Mass., 1987.

\bibitem[Latour and Woolgar(1979)]{Latour+Woolgar:79a}
Bruno Latour and Steve Woolgar.
\newblock \emph{Laboratory life: The social construction of scientific facts}.
\newblock Sage Publications, Beverly Hills, 1979.

\bibitem[Lewin(1998)]{Lewin:98a}
Beverly~A. Lewin.
\newblock {Hedging: Form and function in scientific research texts}.
\newblock In \emph{Genre Studies in English for Academic Purposes}, volume~9,
  pages 89--108. Universitat Jaume I, 1998.

\bibitem[Lin et~al.(2006)Lin, Wilson, Wiebe, and Hauptmann]{Lin+al:06a}
Wei-Hao Lin, Theresa Wilson, Janyce Wiebe, and Alexander Hauptmann.
\newblock Which side are you on? identifying perspectives at the document and
  sentence levels.
\newblock In \emph{CoNLL}, 2006.

\bibitem[Myers(1989)]{Myers:89a}
Greg Myers.
\newblock {The pragmatics of politeness in scientific articles}.
\newblock \emph{Applied Linguistics}, 10\penalty0 (1):\penalty0 1--35, 1989.

\bibitem[Myers(2003)]{Myers:2003a}
Greg Myers.
\newblock {Discourse studies of scientific popularization: Questioning the
  boundaries}.
\newblock \emph{Discourse Studies}, 5\penalty0 (2):\penalty0 265--279, 2003.

\bibitem[Salager-Meyer(2011)]{Salager-Meyer:2011a}
Fran\c{c}oise Salager-Meyer.
\newblock {Scientific discourse and contrastive linguistics: hedging}.
\newblock \emph{European Science Editing}, 37\penalty0 (2):\penalty0 35--37,
  2011.

\bibitem[Scheufele(1999)]{Scheufele:99a}
Dietram~A. Scheufele.
\newblock {Framing as a theory of media effects}.
\newblock \emph{Journal of Communication}, 49\penalty0 (1):\penalty0 103--122,
  1999.

\bibitem[Vapnik(1998)]{Vapnik:98}
Vladimir~N. Vapnik.
\newblock \emph{Statistical Learning Theory}.
\newblock Wiley-Interscience, 1998.

\bibitem[Varttala(1999)]{Varttala:99a}
Teppo Varttala.
\newblock {Remarks on the communicative functions of hedging in popular
  scientific and specialist research articles on medicine}.
\newblock \emph{English for Specific Purposes}, 18\penalty0 (2):\penalty0
  177--200, 1999.

\bibitem[Varttala(2001)]{Varttala:2001a}
Teppo Varttala.
\newblock \emph{Hedging in scientifically oriented discourse: Exploring
  variation according to discipline and intended audience}.
\newblock PhD thesis, University of Tampere, 2001.

\end{thebibliography}
\end{document}